\documentclass{article}

\usepackage{arxiv}

\usepackage[utf8]{inputenc} % allow utf-8 input
\usepackage[T1]{fontenc}    % use 8-bit T1 fonts
\usepackage{hyperref}       % hyperlinks
\usepackage{url}            % simple URL typesetting
\usepackage{booktabs}       % professional-quality tables
\usepackage{amsfonts}       % blackboard math symbols
\usepackage{nicefrac}       % compact symbols for 1/2, etc.
\usepackage{microtype}      % microtypography
\usepackage{lipsum}

\usepackage{graphicx}
\usepackage{comment}
\usepackage{amsmath,amssymb} % define this before the line numbering.
\usepackage{color}
\usepackage{bbm}
\usepackage{multirow}
\usepackage{subcaption}
\usepackage[flushleft]{threeparttable}

\title{MAGNet: Multi-Region Attention-Assisted Grounding of Natural Language Queries at Phrase Level}

\author{
  Amar Shrestha \thanks{equal contribution}\\
  Syracuse University \\
  \texttt{amshrest@syr.edu} \\
   \And
 Krittaphat Pugdeethosapol \footnotemark[1]\\
  Syracuse University \\
  \texttt{kpugdeet@syr.edu} \\
   \And
 Haowen Fang \\
  Syracuse University \\
  \texttt{hfang02@syr.edu} \\
   \And
 Qinru Qiu \\
  Syracuse University \\
  \texttt{qiqiu@syr.edu} \\
}

\date{}

\hypersetup{
pdftitle={MAGNet: Multi-Region Attention-Assisted Grounding of Natural Language Queries at Phrase Level},
pdfauthor={Amar Shrestha, Krittaphat Pugdeethosapol, Haowen Fang, Qinru Qiu},
pdfkeywords={Visual Grounding, Referring Expression, Phrase Localization},
}

\begin{document}

\maketitle

\begin{abstract}
Grounding free-form textual queries necessitates an understanding of these textual phrases and its relation to the visual cues to reliably reason about the described locations. Spatial attention networks are known to learn this relationship and focus its gaze on salient objects in the image. Thus, we propose to utilize spatial attention networks for image-level visual-textual fusion preserving local (word) and global (phrase) information to refine region proposals with an in-network Region Proposal Network (RPN) and detect single or multiple regions for a phrase query. We focus only on the phrase query - ground truth pair (referring expression) for a model independent of the constraints of the datasets i.e. additional attributes, context etc. For such referring expression dataset ReferIt game, our Multi-region Attention-assisted Grounding network (MAGNet) achieves over  12\% improvement over the state-of-the-art. Without the context from image captions and attribute information in Flickr30k Entities, we still achieve competitive results compared to the state-of-the-art. 

% Without utilizing additional information, our Multi-region Attention-assisted Grounding network (MAGNet) achieves respectable results   in Flickr30k entities and over 12\% improvement over the state-of-the-art in ReferIt game dataset. Utilizing in-network RPN and attention also provides a self-sufficient model and interpretable results respectively.
\end{abstract}

\section{Introduction} \label{sec:intro}

Object detection has been the bread and butter of computer vision with the recent advances in deep learning leading to super-human performances in terms of accuracy and speed \cite{li2019scale} \cite{liu2019cbnet}. 
% But object detection is dependent upon the fixed number of classes it is trained with, thus, requiring a pre-defined set of classes. And it also doesn’t acquire complete understanding of the classes as it is unaware of their context and description. This essentially means the object detectors cannot detect objects/regions of interest in the image based on natural language descriptions.
A variation of the object detection task is visual grounding where the objective is to detect objects/regions of interests in the image referenced by a descriptive phrase instead of a pre-defined set of classes. The visual grounding task can have various specific objectives: (a) Phrase localization \cite{plummer2015flickr30k} \cite{wang2018learning}: The language query is a local phrase from a caption describing an image such that  an image region linked to the phrase may or may not be independent of the broader context of the full caption. This makes the queries inherently ambiguous. (b) Referring expression \cite{kazemzadeh2014referitgame} \cite{Mao2016} \cite{yu2016modeling} \cite{yu2018mattnet} \cite{liu2019improving}: the query is an expression referring to a particular region of an image. It is less ambiguous. (c) Natural language object retrieval \cite{hu2016natural} \cite{li2017deep}: a query is used to retrieve images from a set of images. (d) Visual question answering  \cite{gan2017vqs} \cite{li2018tell} \cite{zhu2016visual7w}: a query is in the form of a question and the image region is the associated answer.

In this paper, we will mainly focus on the tasks (a) and (b). 
% To solve these tasks, the framework needs to understand the query and the image and be able to relate between those modes. Given an image, a region may be described by multiple queries or a single query may describe multiple regions in the image. Thus, making the task inherently ambiguous. 
Recently, various approaches have been developed to solve the above-mentioned specific tasks. Most state-of-art visual grounding systems have a two-stage framework \cite{wang2018learning} \cite{yu2016modeling} \cite{yu2018mattnet} \cite{chen2017query} \cite{deng2018visual} \cite{plummer2018conditional} \cite{mao2016generation} which rely on an explicit pre-trained object detector to obtain proposed object bounding boxes and rank their ROI-pooled features based on the encoded feature obtained from the query. This essentially limits these systems to a fixed set of object classes that the detector was trained on. One-stage approaches \cite{yeh2017interpretable} \cite{chen2018real} \cite{sadhu2019zero} \cite{yang2019fast} adopt object detection frameworks to generate image features of all possible regions and fuse them with separately encoded features for the query (proposal-level visual-textual fusion) to rank them. Such proposal-level fusion doesn't build an understanding of the whole image in relation to the phrase query. Some datasets \cite{plummer2015flickr30k} also provide annotations in addition to the query such as class, attribute, etc. described by the query and thus are used in various works \cite{yu2018mattnet} \cite{chen2018real} \cite{liu2017referring}. This makes them dependent on the information provided by the dataset and not purely based on the natural language query. To reduce the ambiguity in phrase localization, [24] also utilizes the full sentence to describe the image along with the query to develop relationships between multiple queries in the sentence.

Evaluation metrics used to measure the performance also adds bias to some existing works. The conventional $Recall@K$ metric essentially expects the predicted region in an image to be ranked in the top K spots. Thus, most works are designed to predict one region per query even if the query might suggest multiple regions in the image irrespective of how the dataset has marked the ground truth.

In this work, to address the mentioned issues, we utilize an encoder-decoder language model with spatial attention for image-level visual-textual fusion of the input image and the natural language query which encodes both the local (word) and global (full query/phrase) understanding of the query in relation to the input image. We utilize this context generated from the attention distribution to train a Faster-RCNN framework \cite{ren2015faster} such that the proposal generation through in-network Region Proposal Network (RPN) is trained to understand the multi-modal relationship and is not limited to a fixed set of classes, and the Region-CNN network is trained to detect one or multiple regions that can relate to the given query. We depend only on the phrase query - ground truth pair information to make the model independent of the constraints of the datasets i.e. additional attributes, context etc. We call this framework Multi-region Attention-assisted Grounding network (MAGNet). We evaluate our approach on Flickr30k entities \cite{plummer2015flickr30k}, ReferIt game \cite{kazemzadeh2014referitgame} and Visual Genome \cite{krishna2017visual} datasets. Thus, the contributions of this work are listed as follows:

\begin{itemize}
\item Image-level visual-textual fusion of the input image and the natural language query through the encoder-decoder language model with spatial attention.
\item Spatial Attention distribution representing global (phrase) understanding alongside the local (word) understanding of the query in relation to the input image.
\item Attention-assisted proposal generation through in-network RPN trained on the context generated from attention.
\item Attention-assisted region detection through Region-CNN trained on the context generated from attention enabling single or multiple detections for a single query.
\end{itemize}

\section{Related works} \label{sec:related_works}

\begin{figure}[t]
\centering
\begin{subfigure}{0.8\textwidth}
\centering
\includegraphics[height=6.0cm]{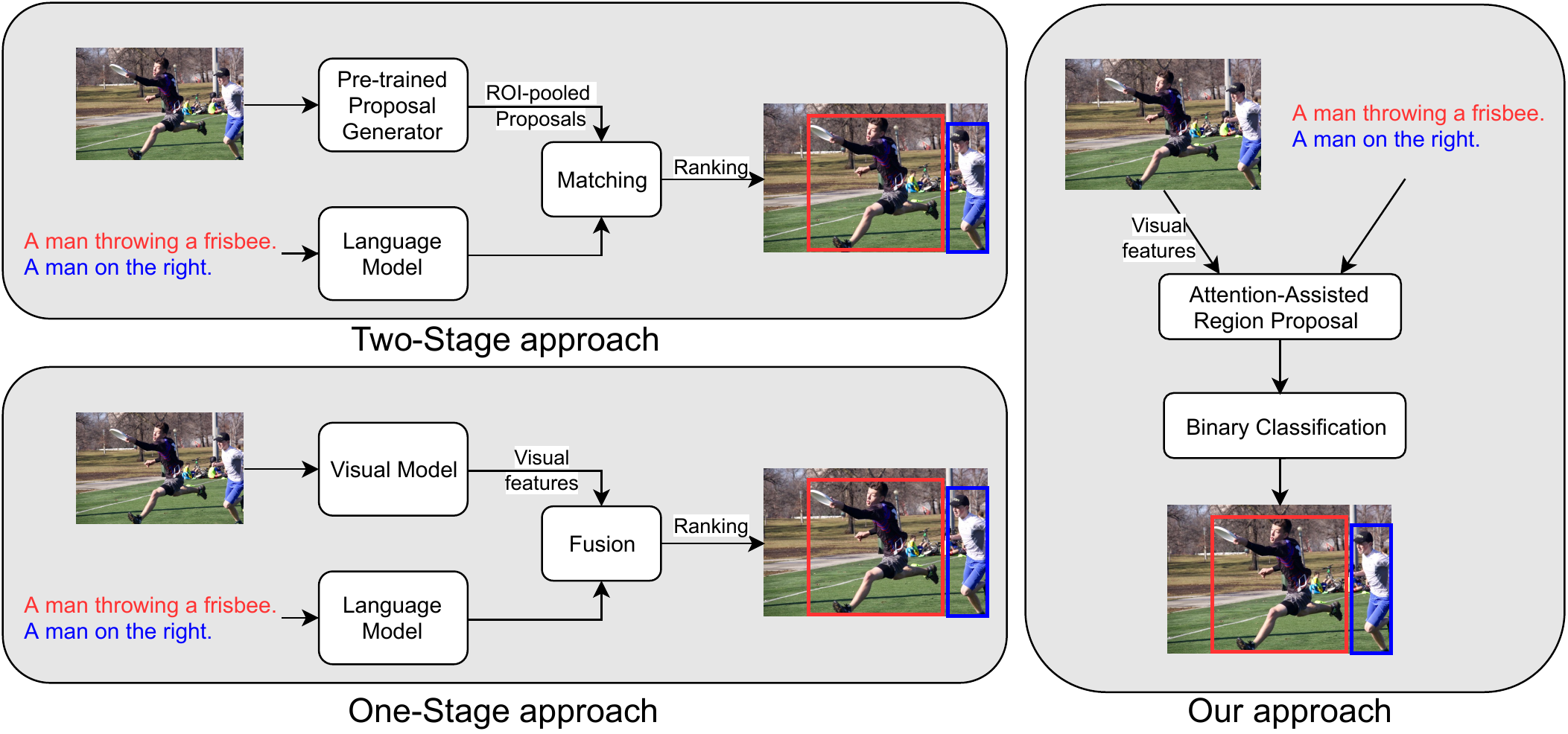}
\caption{}
\label{fig:approach_compare}
\end{subfigure}
\centering
\begin{subfigure}{0.8\textwidth}
\centering
\includegraphics[height=5.0cm]{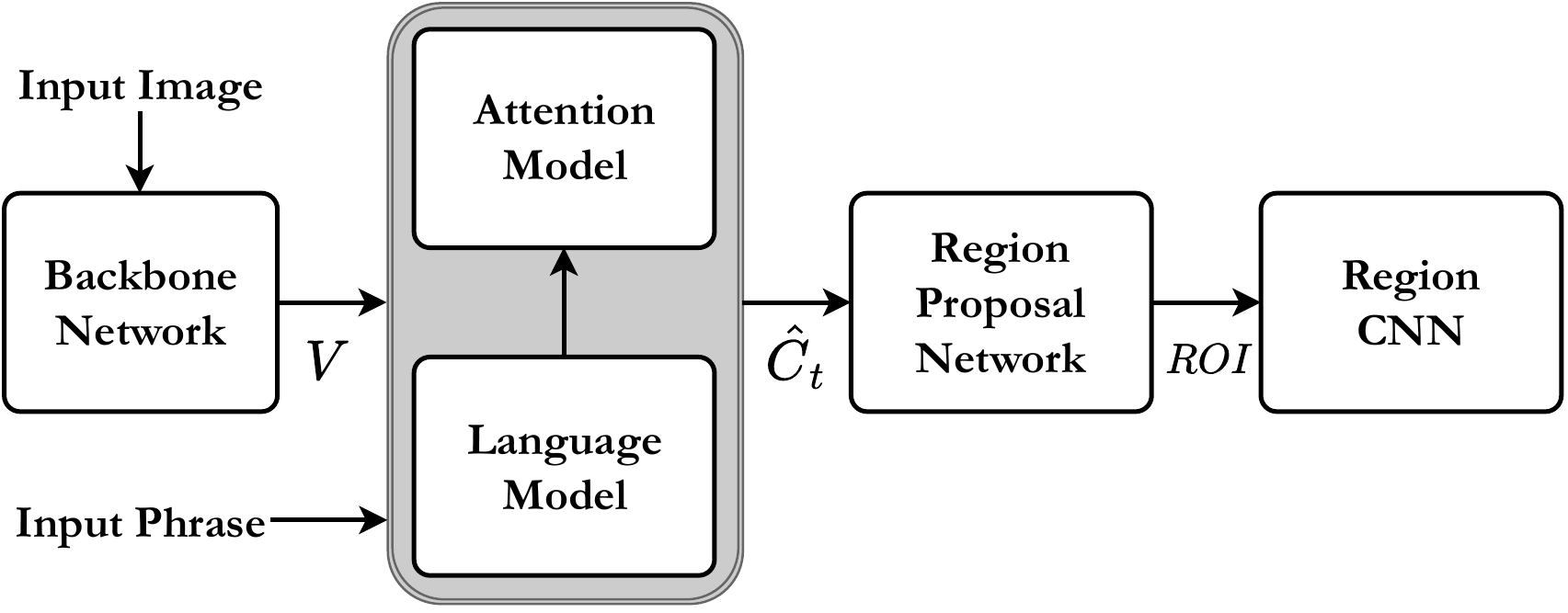}
\caption{}
\label{fig:block_diagram}
\end{subfigure}
\caption{(a) Visual grounding approaches (b) Block diagram of our model. }
\label{fig:image2}
\end{figure}

As we intend to focus on the phrase localization and referring expression tasks in a supervised setting, we compare our work to related works specifically for those tasks. Fig. \ref{fig:approach_compare} shows the types of appraoches.

\textbf{Two-stage approach.} The majority of the grounding systems follow a two-stage approach: proposal generation and ranking. Proposal generation is performed either through a pre-trained RPN \cite{chen2017query} \cite{kovvuri2018pirc} or Faster-RCNN  \cite{yu2018mattnet} \cite{liu2017referring}, proposal generation algorithms such as Edgebox \cite{plummer2018conditional} \cite{wang2018learning}, Multibox  \cite{mao2016generation}, Selective Search  \cite{rohrbach2016grounding} or proposal candidates based on all the ground truths in the image  \cite{deng2018visual} \cite{liu2017referring}. The proposals are then matched with an encoding of the query and then ranked using ranking algorithm or network based on their matching scores. The performance of these two-stage systems relies heavily on the proposal generation. And as the proposal generation mostly focuses on just objects when it's trained as object detectors, the regions unrelated to objects are often missed. For example, generated proposals may contain “person”, “tree”, “car” etc. but it might not contain “sky to the left of the tree”, “a group of people”, etc.

\textbf{One-stage approach.}  \cite{yeh2017interpretable} \cite{chen2018real} \cite{sadhu2019zero} \cite{yang2019fast} adopt object detection frameworks such as to SSD  \cite{liu2016ssd}, YOLO \cite{redmon2016you}, FPN  \cite{lin2017feature} and Retina Net \cite{lin2017focal} to generate image features of all possible regions and fuse them with separately encoded features for the query to rank them. As an encoding does not capture the entire information, fusing them after encoding might lead to loss of relationships between the two modes (i.e. image and query). 

\textbf{Additional information for reducing ambiguity.} Various works also utilize additional information other than the image and query-ground truth pair to reduce ambiguity and fine-tune the grounding predictions. Some datasets  \cite{plummer2015flickr30k} provide annotations in addition to the query such as class, attribute, etc. described by the query. As some of the two-stage approaches \cite{yu2018mattnet}  \cite{chen2018real}, one stage approaches \cite{chen2018real} \cite{liu2017referring} also utilizes the attribute classes to refine the grounding. The help of these attribute is noticeable in the performance but are not available for most visual grounding datasets. In phrase localization task, the image caption is available. This helps reduce the ambiguity of just utilizing the query phrase. \cite{kovvuri2018pirc} \cite{chen2017query} utilize the image caption to form relations between the query phrases to improve grounding performance. But, datasets for referring expressions do not have such captions relating query phrases in the image. And most of the works including one-stage approaches \cite{yeh2017interpretable} \cite{sadhu2019zero} \cite{yang2019fast} also encode the spatial information as an 8-dimensional feature vector to bias predictions for queries based on their location.

\textbf{One query – one region approach.} The majority of the related works are designed to output only one region for a query. This bias is derived from the current formalization of the visual grounding problem and prevalent use of Recall@K metric to evaluate the performance. This metric essentially expects the predicted region in an image to be ranked in the top K spots. Thus, most work either utilize matching network/algorithms to generate a matching score to produce ranking for proposed regions or simply utilize a softmax over the proposed regions. These systems are thus unable to localize a query to multiple regions in the image even though multiple objects matching the query exists.

\textbf{Our approach.} In this work, we intend to enable visual grounding for single or multiple regions in an image based on natural language query without the use of any additional information other than the image and query-ground truths pairs and any pre-trained proposal generation systems. The approach and evaluation of the approach are described in the following sections.

\section{Methodology} \label{sec:methodology}

In this section, we describe the MAGNet framework. 
% We propose an adaptation to the Region Proposal Network in the Faster-RCNN  \cite{ren2015faster} setting such that the resulting architecture is suitable for localizing free-form phrase queries to single or multiple regions in the image. 
Our approach involves encoding the image and phrase using an Encoder-Decoder framework (Section \ref{sec:encoder_decoder}), identifying regions of interest using a spatial attention model (Section \ref{sec:attention}) embracing both local and global information and integrating the attentions into a region proposal network (Section \ref{sec:attention_rpn}) and region-CNN  (Section \ref{sec:attention_cnn}). In the following sections, we introduce our model. In Section \ref{sec:ablation} we perform an ablation study. The block diagram of the overall framework is shown in Fig. \ref{fig:block_diagram}.

\begin{figure}[t]
\centering

\begin{subfigure}{1.0\textwidth}
\centering
\includegraphics[height=8.0cm]{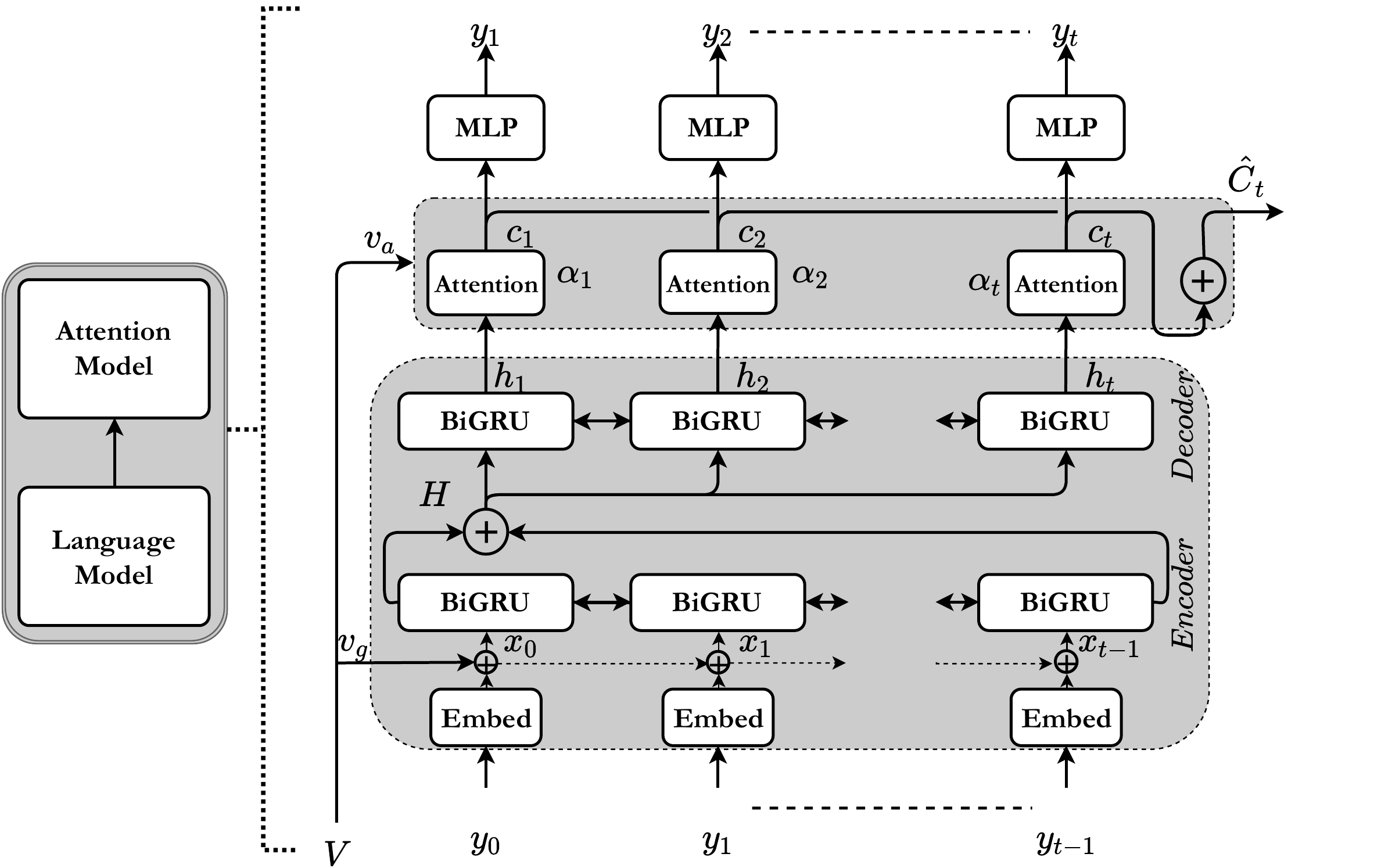}
\caption{}
\label{fig:encoder_decoder}
\end{subfigure}
% \begin{subfigure}{1.0\textwidth}
\begin{subfigure}{1.0\textwidth}
\centering
\includegraphics[height=3.7cm]{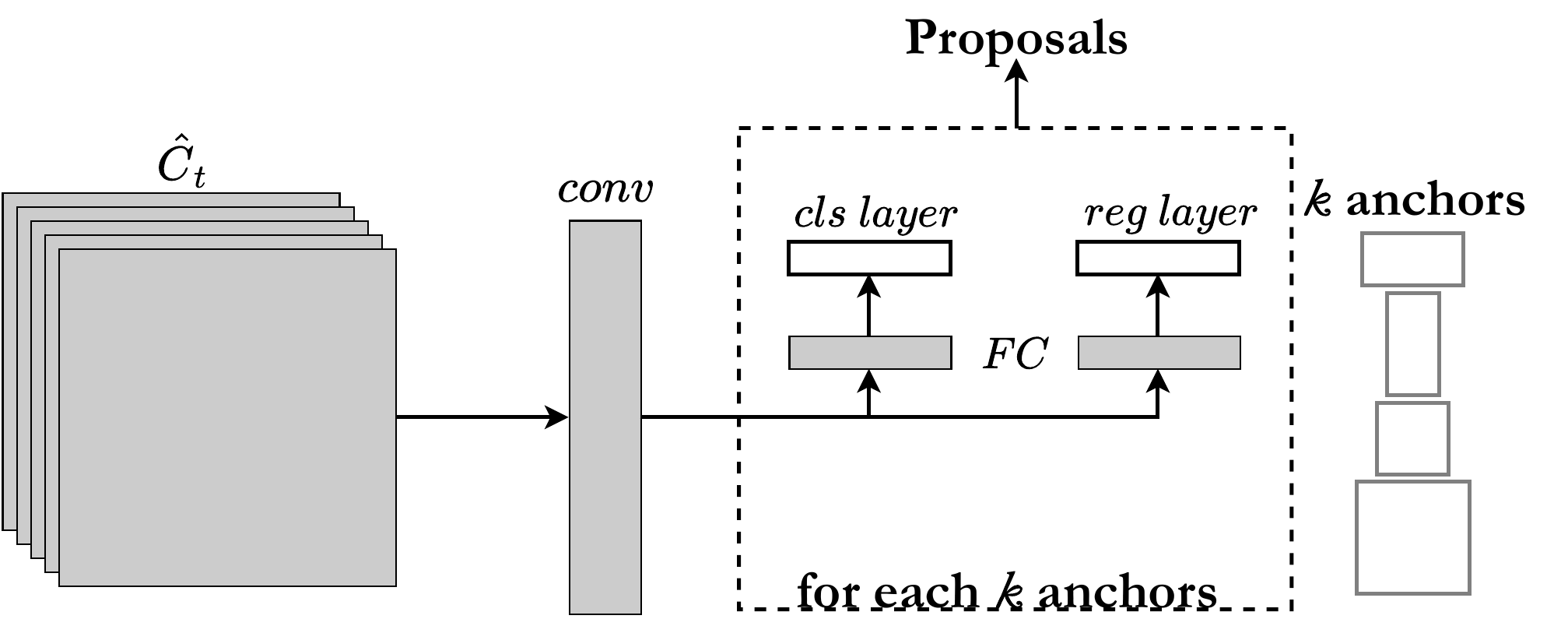}
\caption{}
\label{fig:attention_rpn}
\end{subfigure}

\begin{subfigure}{1.0\textwidth}
\centering
\includegraphics[height=4.6cm]{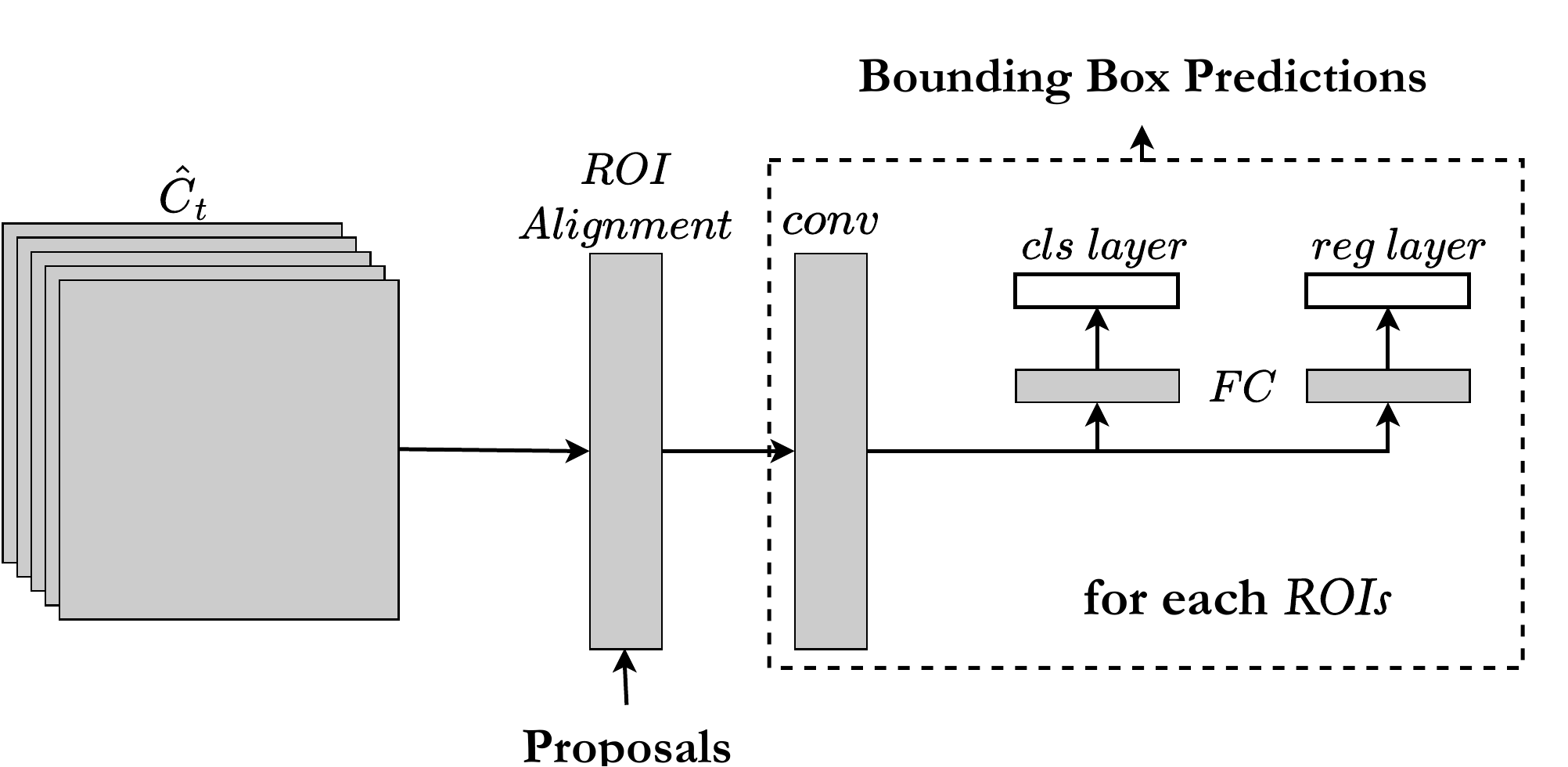}
\caption{}
\label{fig:attention_cnn}
\end{subfigure}
% \end{subfigure}

\caption{(a) Encoder-Decoder Language Model. (b) Attention-based Region Proposal Network. (c) Attention-based Region CNN. }
\label{fig:image3}
\end{figure}

\subsection{Visual features} \label{sec:visual_features}

For our model, we use ResNet-50 \cite{babenko2015aggregating} as the backbone network to extract visual features of the input image. The input image is resized and padded with zeros to get a square image of size $512 \times 512$. The visual features are extracted from the $C4$ layer with size $32 \times 32 \times D_c$ such that the feature map is $1/16$ of the input image and $D_c$ is the number of feature maps. The choice of $C5$ $(16 \times 16)$ and $C4$ $(32 \times 32)$ makes minimal difference in the performance. $D_c$ varies with the choice of the backbone and the input image size. So, to standardize the model, we add a $1 \times 1$ convolutional layer with a ReLU and $D_e=512$ filters to produce the final visual features $V$. This visual feature is further encoded separately for the language model and the spatial attention model.

\subsection{Encoder-Decoder Language Model} \label{sec:encoder_decoder}

To encode the image and the phrase together, we first adopt the encoder-decoder framework \cite{bahdanau2014neural} and modify it to encode the image and corresponding phrase together. Fig. \ref{fig:encoder_decoder} shows the encoder-decoder model we utilize.
% Such that the encoder-decoder framework directly maximizes the following objective:

% % \vspace{-0.3cm}
% \begin{equation} \label{eq:objective}
%     \theta^* = \underset{\theta}{\text{arg max}}\sum_{(I,y)}\log p(y|I;\theta)
% \end{equation}
% % \vspace{-0.3cm}

% where $\theta$ are the parameters of the model, $I$ is the image and $y = \{y_1,...,y_{t-1}\}$ is the corresponding phrase such that 

% % \begin{equation} \label{eq:p_y_i}
% %     \log P(y|I) = \sum_{t=1}^T \log p(y_t|y_1,\ldots,y_{t-1},I)
% % \end{equation}

% % \begin{equation} \label{eq:p_y_y}
% %     \log P(y_t|y_1,\ldots,y_{t-1},I) = f(h_t, c_t)
% % \end{equation}

% % \vspace{-0.5cm}
% \begin{gather}
% \log P(y|I) = \sum_{t=1}^T \log p(y_t|y_1,\ldots,y_{t-1},I) \label{eq:p_y_i} \\
% \log P(y_t|y_1,\ldots,y_{t-1},I) = f(h_t, c_t) \label{eq:p_y_y}
% \end{gather}
% % \vspace{-0.5cm}

% where $c_t$ is the visual context vector, $h_t$ is the hidden state at time $t$ and $f$ is a nonlinear function represented by a recurrent neural network (RNN) that outputs the probability of $y_t$, then 

% % \vspace{-0.3cm}
% \begin{equation} \label{eq:ht_rnn}
%     h_t = RNN(x_t, h_{t-1})
% \end{equation}
% % \vspace{-0.5cm}

% where $x_t$ is the input vector combining the phrase and image information. 

% Fig. \ref{fig:encoder_decoder} shows the encoder-decoder model we utilize based on the framework given by equations \ref{eq:p_y_i} \ref{eq:p_y_y} and \ref{eq:ht_rnn}. 
In this work, for our model, we adopt Gated Recurrent Units (GRU) instead of Long-Short Term Memory (LSTM) as it has demonstrated state-of-the-art performances with significantly lower number of parameters. Alongside, we adopt a bidirectional version of GRUs to encode the phrases from both front-to-back and back-to-front. The encoder encodes the combination of the image and phrase producing the global embedding $H$. 
% \vspace{-0.3cm}
\begin{equation} \label{eq:encbigru}
    H = Encoder(x_t, eh_{t-1}, m_{t-1}, eh_{t+1}, m_{t+1})
\end{equation}
% \vspace{-0.3cm}

$m_t$ is the memory cell vector and $eh_t$ is the hidden state of the encoder at the time before and after $t$. $x_t$ is the input vector formed from the concatenation of the visual features from the image and embedding of the words $y_t$ in the phrase at each time t given as $E_t$. For our model, the visual features from the backbone network V are embedded using global average pooling (GAP) such that $v_g=GAP(V)$. The phrase is embedded using pre-trained 300-dimensional GLOVE embedding vectors \cite{pennington2014glove} $w_t$ trained on Wikipedia2014 and Gigaword. Such that $x_t = [E_t;v_g]$

% % \vspace{-0.3cm}
% \begin{equation} \label{eq:x_t}
%     x_t = [E_t;v_g]
% \end{equation}
% % \vspace{-0.5cm}

The encoder may consist of multiple recurrent layers, but for our model, we only use a single layer of BiGRU. The decoder also only consists of a single layer of BiGRU such that its hidden state is represented as:

% \vspace{-0.3cm}
\begin{equation} \label{eq:decbigru}
    h_t = Decoder(H, h_{t-1}, m_{t-1}, h_{t+1}, m_{t+1})
\end{equation}
% \vspace{-0.3cm}

Utilizing the context vector $c_t$ generated from attention distribution to be detailed in Section \ref{sec:attention} and hidden vector $h_t$, the probability distribution of $y_t$ over the word vocabulary is generated as:

% \vspace{-0.3cm}
\begin{equation} \label{eq:p_yt_i_encdec}
    p (y_t | I) = f(h_t, c_t)
\end{equation}
% \vspace{-0.5cm}

\subsection{Attention model} \label{sec:attention}

%The visual context vector $c_t$ at time $t$ in equation \ref{eq:p_y_y} from image $I$ can be modelled in various ways. 
In the attention-based frameworks such as \cite{xu2015show} at time $t$ based on the hidden state, the decoder would focus on the specific regions of the image with a distribution $a_t$ and compute $c_t$ using the spatial image features from visual backbone network. Such that the context vector $c_t$ is defined as:

% \vspace{-0.3cm}
\begin{equation} \label{eq:c_t}
    c_t = g(v_a, h_t)
\end{equation}
% \vspace{-0.5cm}

where $g$ is the attention function that will be given later by equation \ref{eq:c_t_alpha_t_v_t}, and  $v_a=Conv_{1 \times 1}(V)$ with number of filters $D_a$ to match dimensions with $h_t$. $v_a \in \mathbb{R}^{D_a \times D_f}$ where $D_f=w_f \times h_f$ is the number of pixels in a single visual feature map.  For a $512 \times 512$ input image, $D_f$ is $32 \times 32$ and each pixel in the feature map corresponds to a $16 \times 16$ region in the input image. 

$c_t$ in equation \ref{eq:c_t} captures the region of focus in the visual features pertaining to the current word $t$ in the phrase. In localizing the phrase, it is important to preserve the global information of the whole phrase when focusing on a region. $H$ encodes the entire phrase in the encoder layer such that it is a viable candidate to generate the context vector. The importance of having $H$ will be shown in Section \ref{sec:ablation}. With the global information $H$, the context vector $c_t$ can be derived by the following: 

% \vspace{-0.3cm}
\begin{equation} \label{eq:c_t_with_h}
    c_t = g(v_a, h_t, H)
\end{equation}
% \vspace{-0.5cm}

where $H \in \mathbb{R}^{D_a \times 1}$. Given $v_a$, $h_t$ and $H$, we apply a simple neural network and a softmax function to generate the attention distribution $\alpha_t$ over the spatial image features at time $t$:

% \begin{equation} \label{eq:z_t}
%     z_t = W_z^T \tanh{(W_v v_a + (W_h h_t) \mathbbm{1}^T + (W_H H)\mathbbm{1}^T)}
% \end{equation}

% \begin{equation} \label{eq:alpha_t}
%     \alpha_t =\text{softmax}(z_t)
% \end{equation}

% \vspace{-0.5cm}
\begin{gather}
z_t = W_z^T \tanh{(W_v v_a + (W_h h_t) \mathbbm{1}^T + (W_H H)\mathbbm{1}^T)}  \label{eq:z_t} \\
\alpha_t =\text{softmax}(z_t) \label{eq:alpha_t}
\end{gather}  
% \vspace{-0.5cm}

where $W_v, W_h, W_H \in \mathbb{R}^{D_f \times D_a}$, and $W_z \in \mathbb{R}^{D_f \times 1}$ are weight coefficients learned from the training process, and $\mathbbm{1} \in \mathbb{R}^{D_f \times 1}$ such that $a_t \in \mathbb{R}^{1 \times D_f}$. The context vector $c_t$ at time $t$ can now be obtained as:

% \vspace{-0.3cm}
\begin{equation} \label{eq:c_t_alpha_t_v_t}
    c_t = v_a (W_H H) \alpha_t
\end{equation}
% \vspace{-0.5cm}

Such that we model the probability distribution over $y_t$ in equation \ref{eq:p_yt_i_encdec} as

% \vspace{-0.3cm}
\begin{equation} \label{eq:p_yt_i}
    p (y_t | I) = f(h_t, c_t) = \text{softmax}(W_p(c_t + h_t \mathbbm{1}^T))
\end{equation}
% \vspace{-0.5cm}

where $W_p$ is learned weight matrix. The log probability distribution $y_t$ is maximized with a cross-entropy loss. Applying this loss as an auxiliary loss enables training the attention vector without any grounding supervision.

\subsection{Attention-based Region Proposal Network} \label{sec:attention_rpn}

Instead of using a pre-trained RPN to generate proposals in conventional two-stage phrase localization works, we intend to train the RPN with assistance from the context $(c_t)$ derived from encoding the visual and phrase features together. 

The context vector $c_t$ from equation \ref{eq:c_t_alpha_t_v_t} represents the understanding of the word in a phrase at time t in terms of focus on the image. To utilize this context, we need to combine the context over the entire phrase. In our model, we simply average the context over the time dimension such that the resulting context $\hat{C}_T$ still has the same dimensions as the original visual feature $v_a$.

% \vspace{-0.3cm}
\begin{equation} \label{eq:c_hat}
    \hat{C}_T = \frac{1}{T} \sum_{t=1}^T c_t
\end{equation}
% \vspace{-0.5cm}

where $T$ represents the number of words in the phrase.

We use this average context vector $\hat{C}_T$ as the input of the RPN. Similar to Faster-RCNN, our RPN as shown in Fig. \ref{fig:attention_rpn} takes the average context vector as input and outputs a set of rectangular object proposals ($reg$ layer), each with an objectness score ($cls$ layer). Here, we define "objectness" not literally but based on how the phrases are grounded in the dataset. For example, the phrase "a red shirt" refers to an actual object whereas the phrase "a group of people" might not fit the literal definition of the word “object” but still is taken as such based on the dataset.

We also adopt the same multi-task loss as in \cite{ren2015faster} to train the $cls$ layer (binary classification) and the reg layer (regression). In Section \ref{sec:ablation}, we demonstrate the efficacy of utilizing the learned context for training the RPN instead of using a pre-trained RPN.

\subsection{Attention-based Region CNN} \label{sec:attention_cnn}
Now we utilize the proposals generated by RPN for region-based phrase detection CNN. For the detection network, we again adopt Faster-RCNN as shown in Fig. \ref{fig:attention_cnn}. Again, the proposals are used to perform ROI alignment on the context vector $\hat{C}_T$. As we do not have classes as the detection network in Faster-RCNN, we define the task of the $cls$ layer in the phrase detection network as detecting how much the proposal represents the given phrase. For this purpose, $cls$ layer classifies each proposal as either not related or related to the given phrase using a softmax. This essentially means, instead of ranking these proposals, we detect how related these proposals are to the phrase such that we can detect multiple instances of the phrase in the image. The $reg$ layer is now used to regress to the final bounding box for the phrase. After this, we perform a further step of non-maximum suppression to fine-tune the detections.

\section{Experiments} \label{sec:exp}

In this section, we present experiments to evaluate our proposed model MAGNet on varieties of datasets with multiple evaluation metrics and compare our results to the state-of-the-art visual grounding methods  \cite{plummer2015flickr30k} \cite{wang2018learning} \cite{hu2016natural} \cite{chen2017query} \cite{plummer2018conditional} \cite{yeh2017interpretable} \cite{chen2018real} \cite{yang2019fast} \cite{kovvuri2018pirc} and  \cite{rohrbach2016grounding}. Results of ablation studies with different configurations will also be reported to further explain the design decisions of the proposed model.

\subsection{Datasets} \label{sec:datasets}

We evaluate MAGNet on 3 different datasets Flickr30K entities \cite{plummer2015flickr30k}, ReferltGame \cite{kazemzadeh2014referitgame}, and Visual Genome \cite{krishna2017visual}. Flickr30K Entities provides region phrase correspondence annotations to the original Flickr30K. The 31,783 images in Flickr30K have 427K referred entities. We follow the same training/test split used in the previous work \cite{yang2019fast} in our experiments. The queries in Flickr30K are region phrases extracted from a full sentence description of the image. The ground truth image object provided for each query is an object described in the image caption. The contextual information of the image caption imposes extra constraints in visual grounding, such that the dataset ignores other objects in the “background” that also match the query phrase. The MAGNet focuses on the referring expressions itself with no other context information, its training and testing are done solely based on the query phrases. As we will show in this section, it detects more matching objects for the given query. Some of them are not in the ground truth of Flickr30K.
% The region phrases are queries, which are parts of full sentences describing the image. As applying additional context is not in scope of our work, we take phrase queries as independent of the full caption. 
ReferItGame has 20,000 images from the SAIAPR-12 dataset \cite{escalante2010segmented} and contains 130,525 expressions, 96,654 distinct objects, and 19,894 photographs of natural scenes. The queries are expressions referring to one or more regions in the image.  The Visual Genome dataset has a total of 108,077 images with 5.4 million region descriptions. 

\subsection{Evaluation metrics} \label{sec:metrics}

We evaluate the models with two metrics: Recall@K and mean average precision (mAP). Recall@K (R@K) for $K$ = 1, 5 and 10 is defined as the proportion of all positive examples ranked above a given rank K. mAP metric is adapted from the PASCAL VOC challenge \cite{everingham2010pascal} used for object detection tasks. The mAP considers both precision and recall of a model and enables evaluation of the model when there are more than one region to be detected for a single query.
% Recall is defined as the proportion of all positive examples that are predicted correctly. Precision is the proportion of all the predictions that are positive. 
and is defined as the mean precision at a set of eleven equally spaced recall levels $[0, 0.1,... ,1]$. Detailed description of the mAP metric can be found in \cite{everingham2010pascal}. For both metrics, the predicted bounding box is considered positive if it is classified as related to the given query phrase and the intersection over union (IOU) is 0.5 or more.
% $Precision=TP\/(TP+FP)$ and $Recall=TP\/(TP+FN)$ where $TP$, $TN$, $FP$ and $FN$ are true positive, true negative, false positive and false negative respectively. Such that 

% % \vspace{-0.3cm}
% \begin{equation} \label{eq:map}
%     mAP = \frac{1}{11} \sum_{r \in \{0,0.1,...1\}} P_{interp}(r)
% \end{equation}
% % \vspace{-0.5cm}

% where $P_{interp}(r)= \underset{r:\widetilde{r} \geq r}{\max} P(\widetilde{r})$ represents the maximum precision measured for corresponding recall exceeding $r$. 

\subsection{Training details} \label{sec:training}

We reshape the input image to size $512 \times 512$ while keeping the original aspect ratio and padding the smallest dimension with zero pixels. No other data augmentation is performed. Query phrases are prepended with a start token and appended with an end token and embedded with the GLOVE 300D embedding \cite{pennington2014glove}. Shorter phrases are padded with pad tokens and are limited to 18 words. 

We utilize ResNet-50 \cite{he2016deep} trained on ImageNet \cite{russakovsky2015imagenet} as the backbone network for visual features. The two layers of BiGRU contains 512 units each. For the RPN and Region-CNN, we use the same architecture and dimensions as the original Faster-RCNN. For the RPN, we use 9 anchors (3 aspect ratios and 3 sizes) for each feature in the context vector. All the modules in the model are trained together to allow the attention distribution to correlate better with the region proposals and the final region predictions. In the experiments, we found that training the RPN and Region-CNN separately hindered the performance of our approach.

\subsection{Quantitative Analysis} \label{sec:quantitative}

\begin{table}[t]
\begin{center}
\caption{Visual Grounding results}
\label{table:results}
\begin{tabular*}{\textwidth}{l @{\extracolsep{\fill}} lccccccccc}
\hline
 & \textbf{Methods} & \multicolumn{3}{c}{\textbf{Flickr30k Entities}} & \multicolumn{3}{c}{\textbf{ReferltGame}} & \multicolumn{3}{c}{\textbf{Visual Genome}} \\
                          &                & \textbf{R@1}   & \textbf{R@5}   & \textbf{R@10}  & \textbf{R@1}   & \textbf{R@5}   & \textbf{R@10}  & \textbf{R@1}   & \textbf{R@5}   & \textbf{R@10}  \\
\hline
\multirow{10}{*}{\rotatebox{90}{\textbf{2-Stage}}} & SCRC\cite{hu2016natural}           & 27.80 & -     & 62.90 & 17.93 & -     & 45.27 & 11.00 & -     & -     \\
                          & DSPE\cite{wang2016learning}           & 43.89 & 64.46 & 69.66 & -     & -     & -     & -     & -     & -     \\
                          & GroundeR\cite{rohrbach2016grounding}        & 47.81 & -     & -     & 26.93 & -     & -     & -     & -     & -     \\
                          & CCA\cite{plummer2015flickr30k}            & 50.89 & 71.09 & 75.73 & -     & -     & -     & -     & -     & -     \\
                          & Similarity Net\cite{wang2018learning} & 51.05 & 70.30 & 75.04 & -     & -     & -     & -     & -     & -     \\
                          & MSRC\cite{chen2017msrc}           & 57.53 & -     & -     & 32.31 & -     & -     & -     & -     & -     \\
                          & QRN\cite{chen2017query}            & 60.21 & -     & -     & 43.57 & -     & -     & -     & -     & -     \\
                          & QRC\cite{chen2017query}            & 65.14 & -     & -     & 44.07 & -     & -     & -     & -     & -     \\
                          & CITE\cite{plummer2018conditional}           & 61.89 & -     & -     & 34.13 & -     & -     & 24.43 & -     & -     \\
                          & PIRC Net\cite{kovvuri2018pirc}       & \textbf{72.83} & -     & -     & 59.13 & -     & -     & -     & -     & -     \\
\hline
\multirow{4}{*}{\rotatebox{90}{\textbf{1-Stage}}}  & IGOP\cite{yeh2017interpretable}           & 53.97 & -     & -     & 34.70 & -     & -     & -     & -     & -     \\
                          & SSG\cite{chen2018real}            & -     & -     & -     & 54.24 & -     & -     & -     & -     & -     \\
                          & ZSGNet\cite{sadhu2019zero}         & 63.39 & -     & -     & 59.63 & -     & -     & -     & -     & -     \\
                          & \cite{yang2019fast}       & 68.69 & -     & -     & 59.30 & -     & -     & -     & -     & -     \\
                          & \textbf{MAGNet(Ours)}    & 60.20 & \textbf{78.85} & \textbf{79.90} & \textbf{71.60} & \textbf{81.00} & \textbf{81.20} & \textbf{28.85} & \textbf{48.50} & \textbf{50.70} \\
\hline
\end{tabular*}
\end{center}
\end{table}

% \begin{figure}
% \begin{subfigure}{0.99\textwidth}
% \centering
% \includegraphics[height=5cm]{figure/example.pdf}
% \caption{}
% \label{fig:example}
% \end{subfigure}

% \begin{subfigure}{0.99\textwidth}
% \centering
% \includegraphics[height=2.7cm]{figure/positional_grouding.pdf}
% \caption{}
% \label{fig:positional_grounding}
% \end{subfigure}
% \caption{(a) Examples from our approach. (b) Predicted grounding for positional cues.}
% \end{figure}

Table \ref{table:results} compares our approach with prior works on Flickr30k entities, ReferIt, and Visual Genome datasets in terms of Recall@K metric where $K=1,5,$ and 10. We separate the prior works into two-stage and one-stage approaches. and compare the results with our model described in Section \ref{sec:methodology} and \ref{sec:training}. 
% The visual grounding task for Flickr30k entities is phrase localization given the type of dataset and for ReferIt and Visual Genome, it is a referring expression task. 
Results for the prior works are collected from their respective publications. 

\begin{figure}
\begin{subfigure}{0.99\textwidth}
\centering
\includegraphics[height=7cm]{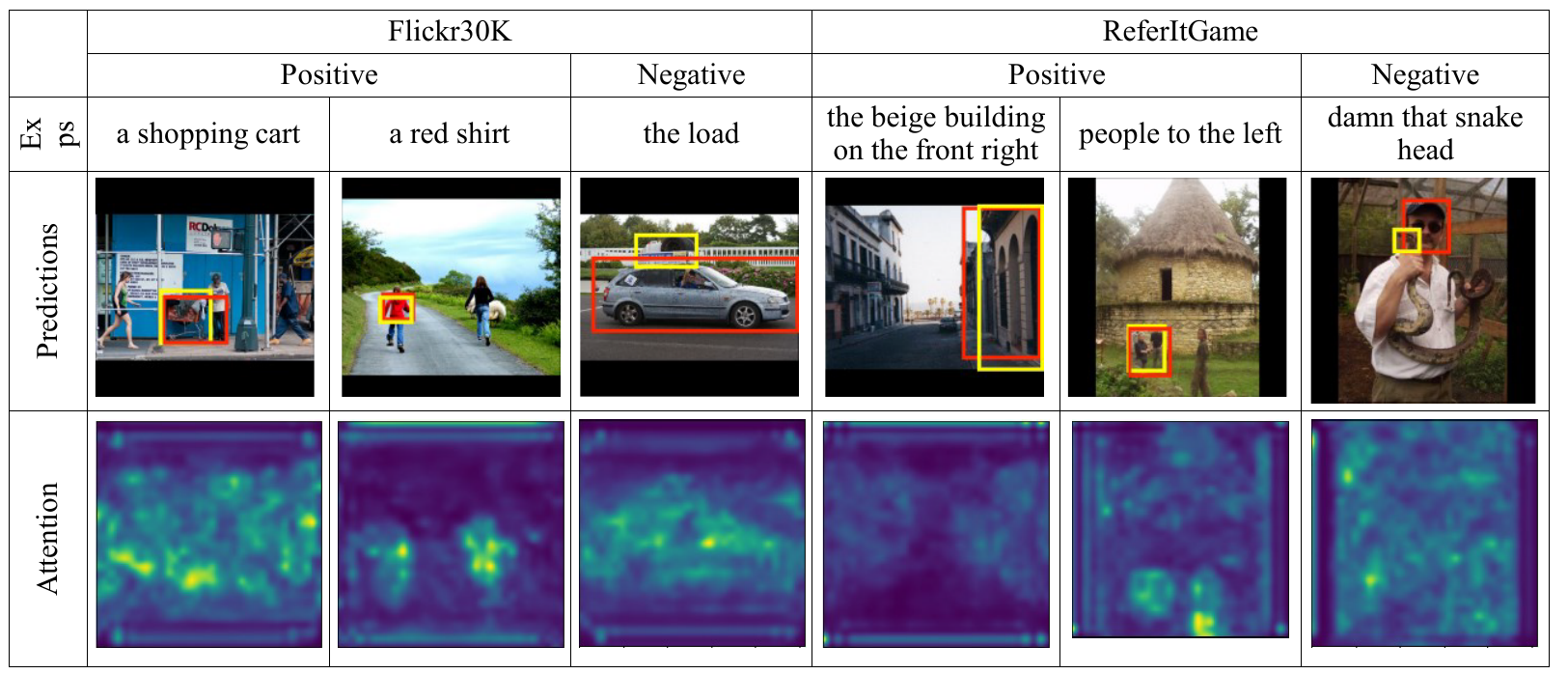}
\caption{}
\label{fig:example}
\end{subfigure}

\begin{subfigure}{0.99\textwidth}
\centering
\includegraphics[height=3.6cm]{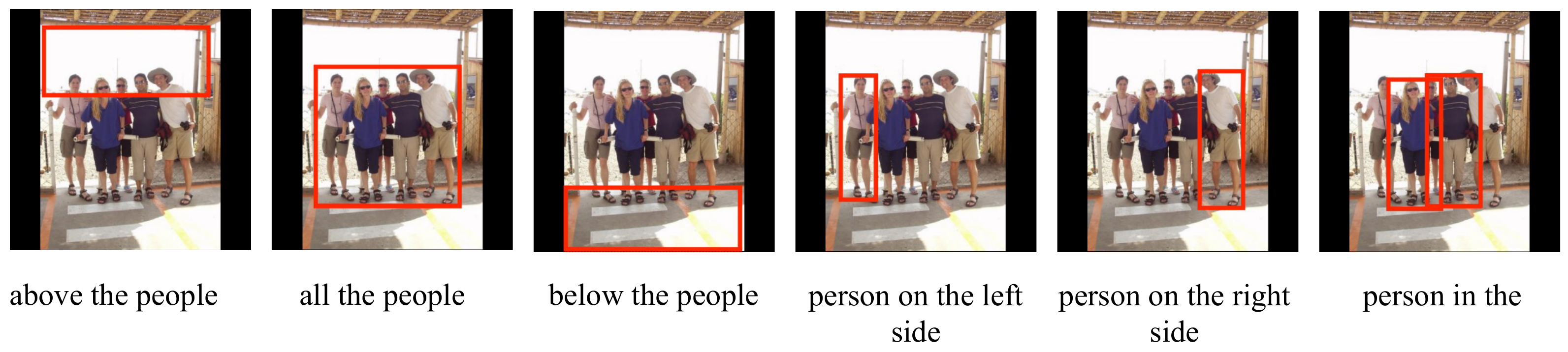}
\caption{}
\label{fig:positional_grounding}
\end{subfigure}

\begin{subfigure}{0.99\textwidth}
\centering
\includegraphics[height=3.2cm]{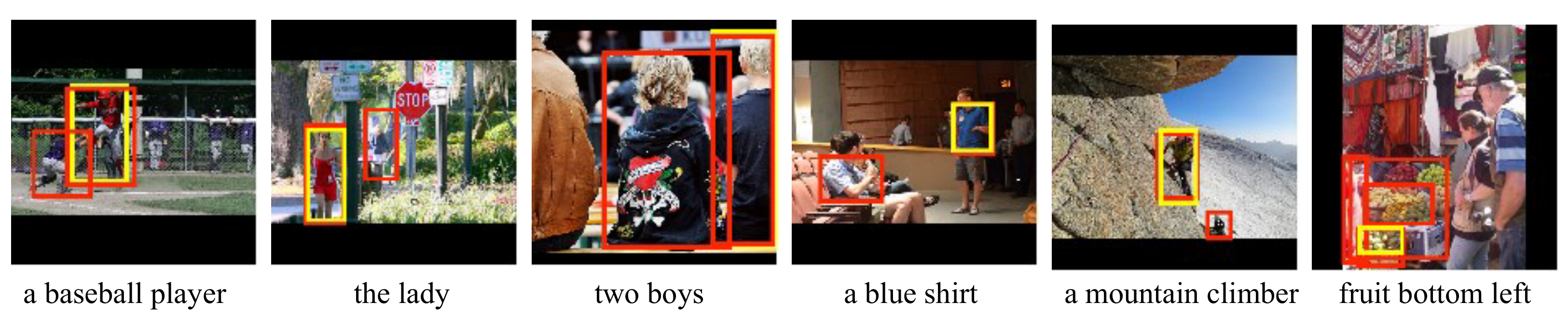}
\caption{}
% \caption{Examples showing multiple region detection and discrepancies with ground truth.}
\label{fig:example_multi_region}
\end{subfigure}

\caption{(a) Examples from our approach. (b) Predicted grounding for positional cues. (c) Examples showing multiple region detection and discrepancies with ground truth.}
\end{figure}

For the phrase localization task on Flickr30k entities, the phrase queries extracted from the image caption ignore the context in the original sentence and thus are highly ambiguous especially in terms of the positional cues of the region. Some examples are given in Figure \ref{fig:example_multi_region} with the ground truths. 
% In these cases, ground truth provided by the dataset is the object mentioned in the image caption. 
As we explained in Section \ref{sec:datasets} , MAGNet searches for the matching objects solely based on the query phrase without considering any additional contextual information. Additionally,  we modelled MAGNet as a detection framework to detect single or multiple regions for a query instead of just specifically one region. Therefore, it is able to detect all matching objects in the image, and the one mentioned in the image caption may not necessarily have the highest score. That is why the R@5 and R@10 score of MAGNet is significantly better than its R@1 score. From this perspective, the Flickr30k entities is not the ideal dataset to evaluate our approach, because only the objects within the context of the image captions are identified as the ground truth, while other objects are ignored even though they also match the query description.
% As applying additional context is not in the scope of our work, we only utilize the natural language query at phrase level, independent of the full caption. Thus, our base model approach lags behind the state-of-the-art methods. 

The authors of \cite{chen2017query} \cite{kovvuri2018pirc} utilize the entire caption to either build relationships between multiple queries in a single image or as the context to reduce the ambiguity in the query phrase. Therefore, they are able to locate the object in the context (i.e. image caption) that matches the phrase description. However, for many applications the caption of the image is usually not available. Furthermore, focusing only on the context given by the image caption prohibits the model to locate all possible matching objects to the query in the image. That is why the performance of these approaches degrade significantly when applied to the ReferItGame dataset. 

% The goal of this work is to localize all objects in the image that matches the phrase description. We modelled our approach as a detection framework to detect single or multiple regions for a query instead of just specifically one region. 
% Thus, for a highly ambiguous task like phrase localization, our approach loses some ground in terms of $R@1$. 
% From this perspective, the Flickr30k entities are not the ideal dataset to evaluate phrase level localization, because only the objects mentioned in the image captions are identified as the ground truth, while other objects are ignored even though they also match the query description. In other words, the ground truth itself does not have 100\% precision for the query at phrase level. Some examples of such insufficiency are given later in Fig. \ref{fig:example} and discussed in Section \ref{sec:qualitative}.

In addition to focus only on a single region, \cite{sadhu2019zero} \cite{yang2019fast} explicitly code spatial features for each position of the spatial dimensions to add positional information thus reducing positional ambiguity. These works also encode the queries and images separately, thus enabling them to utilize powerful pre-trained language models like BERT  \cite{devlin2018bert}. However, separately trained language and image model also means that the model is less effective in extracting language features that have salient image information or vice versa. This probably is another reason that these works perform worse when applied to ReferItGame dataset, where many objects are located based on the position cue in the phrase description.

The advantages of our approach become apparent in the referring expression task in ReferIt and Visual Genome datasets as shown Table \ref{table:results}. In this task, the queries in the image are independent of each other. The queries are self-sufficient with specific positional cues and thus less ambiguous. Hence, the ground truth has better precision in this dataset. Our model achieves 12.30\% better R@1 performance than the current state-of-the-art one-stage  \cite{yang2019fast} and two-stage \cite{kovvuri2018pirc} approaches. This performance boost can be attributed to the following key points: (1). encoding the image and the query together ensures that the query is understood in relation to the given image. Thus, for independent queries, the generated context vector relates the query closely to the image. This is especially effective for queries with positional cues as shown by predicted grounding in Fig. \ref{fig:positional_grounding}. Hence, our framework is better at handling the visual oriented language information. (2) the attention-assisted RPN produces better quality proposals than other pre-trained proposal generators. Table \ref{table:hit} shows the hit rates for various region proposal methods for the number of proposals $N=200$. The MAGNet (a) in the table is the original MAGNet model, however the input of its RPN (Figure \ref{fig:attention_rpn}) is the visual feature $V$ instead of the attention enhanced visual feature ${\hat{C}}_t$. Our attention-assisted RPN produces the highest quality proposals for ReferIt and Visual Genome. 
% Fig. \ref{fig:proposal} shows an example of proposals generated by an (a) RPN trained on MSCOCO, (b) RPN without attention and (c) our attention-assisted RPN. As it can be seen, the RPN (a) produces lots of proposals unrelated and not useful to ground the given phrase, whereas our attention-assisted RPN produces very focused proposals based on the query.

We also achieve state-of-the-art $R@1$ performance on Visual Genome. However, there aren’t many reported visual grounding results on the dataset.

\begin{table}[t]
\begin{center}
\begin{threeparttable}
\caption{Hit rates (N=200) of region proposal methods}
\label{table:hit}
\begin{tabular*}{0.90\textwidth}{l @{\extracolsep{\fill}} ccc}
\hline
\textbf{Method} & \textbf{Flickr30k Entities} & \textbf{ReferIt Game} & \textbf{Visual Genome} \\
\hline
RPN(COCO)\cite{ren2015faster}          & 76.60          & 46.50          & -              \\
Edgebox\cite{zitnick2014edge}            & 83.69          & 68.26          & -              \\
Selective Search\cite{uijlings2013selective}   & 85.68          & 80.34          & -              \\
PGN (N=100)\cite{chen2017query}                & 89.61          & -              & -              \\
% \cite{yang2019fast}                & \textbf{95.48} & 91.32          & -              \\
\textbf{MAGNet}    & \textbf{89.78}          & \textbf{92.68} & \textbf{68.59} \\
\textbf{MAGNet(a)} & 78.22          & 83.98 & 50.90 \\
\hline
\end{tabular*}
\begin{tablenotes}
      \small
      \item MAGNet(a) MAGNet without attention-assisted RPN 
\end{tablenotes}
\end{threeparttable}
\end{center}
\end{table}

\begin{figure*}
    \centering
    \begin{subfigure}{0.3\textwidth}
        \centering
        \includegraphics[height=3.7cm]{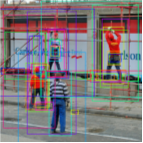}
        \caption{}
        \label{fig:proposal_a}
    \end{subfigure}%
    ~ 
    \begin{subfigure}{0.3\textwidth}
        \centering
        \includegraphics[height=3.7cm]{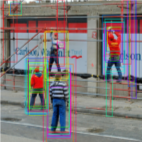}
        \caption{}
        \label{fig:proposal_b}
    \end{subfigure}
    \begin{subfigure}{0.3\textwidth}
        \centering
        \includegraphics[height=3.7cm]{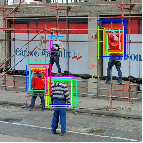}
        \caption{}
        \label{fig:proposal_c}
    \end{subfigure}
    \caption{Proposals generated by query “a red shirt” by (a) RPN(COCO) (b) RPN without attention (c) attention-assisted RPN. The bounding boxes are colored only for distinguishing between the dense boxes}
    \label{fig:proposal}
\end{figure*}

% \begin{figure}
% \centering
% \includegraphics[height=2.4cm]{figure/example_multi_region.pdf}
% \caption{Examples showing multiple region detection and discrepancies with ground truth.}
% \label{fig:example_multi_region}
% \end{figure}

% \begin{table}[]
% \begin{center}
% \begin{threeparttable}
% \caption{Hit rates (N=200) of region proposal methods}
% \label{table:hit}
% \begin{tabular*}{\textwidth}{l @{\extracolsep{\fill}} ccc}
% \hline
% \textbf{Method} & \textbf{Flickr30k Entities} & \textbf{ReferIt Game} & \textbf{Visual Genome} \\
% \hline
% RPN(COCO)\cite{ren2015faster}          & 76.60          & 46.50          & -              \\
% Edgebox\cite{zitnick2014edge}            & 83.69          & 68.26          & -              \\
% Selective Search\cite{uijlings2013selective}   & 85.68          & 80.34          & -              \\
% PGN (N=100)\cite{chen2017query}                & 89.61          & -              & -              \\
% \cite{yang2019fast}                & \textbf{95.48} & 91.32          & -              \\
% \textbf{MAGNet}    & 89.78          & \textbf{92.68} & \textbf{68.59} \\
% \textbf{MAGNet(a)} & 78.22          & \textbf{83.98} & \textbf{50.90} \\
% \hline
% \end{tabular*}
% \begin{tablenotes}
%       \small
%       \item MAGNet: Base model (a) RPN: No attention 
% \end{tablenotes}
% \end{threeparttable}
% \end{center}
% \end{table}

\subsection{Qualitative Analysis} \label{sec:qualitative}

Fig. \ref{fig:example} shows some examples of visual grounding performance of our approach and the attention distribution for Flickr30k entities and ReferIt datasets. For each dataset, the leftmost and middle columns show the visual grounding and attention distribution when the query is grounded correctly, and the rightmost column when the query phrase is grounded incorrectly. The yellow bounding box represents the ground truth and the red bounding box represents the predicted bounding box. As can be seen from the examples, the attention is distributed as suggested by the query focusing on relevant parts as described by the words in the phrase. The incorrect region grounding occurs mainly in cases of high ambiguity in the query. For example, for Flickr30k entities in the rightmost column, the query is “the load” with an image of a car with baggage on its top. The image caption “the passenger is holding on to the load on top of the car” is also provided. The word “load” itself has various meanings. Without the context of the car in the query or the context of the full caption, this example becomes highly ambiguous.

Our approach also detects multiple regions for a query. This causes some discrepancies between our prediction and the ground truth, esp. in Flickr30k entities. We show some examples of discrepancies of grounding of queries in the two datasets, perceived correct grounding and our approach’s predicted grounding in Fig. \ref{fig:example_multi_region}. For example, the query "two boys" is grounded showing only one of the boys in the dataset whereas our approach is able to predict regions for both the boys in the image which is perceived to be correct. 

As mentioned in Section \ref{sec:quantitative}, the attention-assisted RPN produces better quality proposals than other pre-trained proposal generators. Fig. \ref{fig:proposal} shows an example of proposals generated by an (a) RPN trained on MSCOCO, (b) RPN without attention and (c) our attention-assisted RPN. As it can be seen, the RPN (a) produces lots of proposals unrelated and not useful to ground the given phrase, whereas our attention-assisted RPN produces very focused proposals based on the query.

\section{Ablation study} \label{sec:ablation}

We study the effect of some variations in our model to demonstrate the effectiveness of some design choices. Table 3 shows the R@1 and mAP for three variations from the final model.

Variation (a) studies the effect of not using a word embedding. Instead of using the GLOVE 300D embedding, we allow the model to learn the embedding during the training.  This variation has minimal impact on ReferIt whereas a bigger impact on the Flickr30k entities. This is expected as the vocabulary size of ReferIt queries are smaller $(\sim$1500) than that of Flickr30k entities $(\sim4000)$. And also learning the embedding just from the vocabulary doesn't allow the model to generalize as it does when using a word embedding trained on a large corpus.

Variation (b) studies the effect of not utilizing the encoding of the full query $H$ in attention distribution. In this model, (\ref{eq:z_t}) (\ref{eq:alpha_t}) and (\ref{eq:c_t_alpha_t_v_t}) are reduced to the following:

% \vspace{-0.5cm}
\begin{gather}
z_t = w_z^T \tanh (W_v v_a + (W_h h_t)\mathbbm{1}^T ) \label{eq:zt_ablation} \\
a_t = \text{softmax}(z_t) \label{eq:at_ablation} \\
c_t = \alpha_t v_a  \label{eq:alpha_ablation}
\end{gather}  
% \vspace{-0.5cm}

This variation has a clear impact on the performance of our approach for all the datasets. Without the knowledge of the full query, the attention distribution only tends to represent the focus towards the latest word in the query, thus missing the context of the full query. For example, in a query "a red shirt", the attention distribution without $H$ only focuses on shirts at the end of the query, whereas with $H$, the attention is now focusing on red shirt.

Variation (c) studies the effect of training the RPN without the use of the context vector $\hat{C}_T$. In this variation, we directly utilize $v_a$ from equation (10) to train the RPN and utilize $\hat{C}_T$ only to train the Region-CNN. This variation of training RPN is similar to the Proposal Generation Network (PGN) in \cite{chen2017query} but with regular $cls$ and $reg$ RPN loss instead of the proposal generation loss dependent on the context of the full caption. This variation also creates a measurable impact on the performance of our approach. This can be understood as the proposals generated by this RPN are of lower quality than the attention-assisted RPN as shown in Table \ref{table:hit} and Fig. \ref{fig:proposal}.

% \begin{table}[] \footnotesize
% \begin{center}
% \begin{threeparttable}
% \caption{Ablation Study}
% \begin{tabular}{lcccccccccccc}
% \multicolumn{1}{c}{\multirow{2}{*}{Method}} & \multicolumn{4}{c}{Flickr30k Entities} & \multicolumn{4}{c}{Referlt Game} & \multicolumn{4}{c}{Visual Genome} \\
% \multicolumn{1}{c}{} & R@1   & mAP    & R@MC  & P@MC  & R@1   & mAP    & R@MC  & P@MC  & R@1   & mAP    & R@MC  & P@MC  \\
% MAGNet               & 60.20 & 0.4956 & 71.48 & 48.68 & 71.60 & 0.6052 & 79.78 & 66.37 & 28.85 & 0.1892 & 44.46 & 19.60 \\
% MAGNet(a)            & 52.90 & 0.4293 & 67.51 & 44.93 & 69.95 & 0.6129 & 77.65 & 64.20 & 29.00 & 0.1823 & 43.59 & 20.53 \\
% MAGNet(b)            & 49.65 & 0.3672 & 60.73 & 37.63 & 68.00 & 0.5505 & 79.16 & 58.87 & 26.50 & 0.1316 & 41.61 & 16.40 \\
% MAGNet(c)            & 52.90 & 0.3730 & 64.76 & 40.91 & 68.95 & 0.5813 & 77.57 & 62.40 & 28.40 & 0.1144 & 38.8  & 15.46
% \end{tabular}
% \begin{tablenotes}
%       \small
%       \item MAGNet:  model  (a).Word embedding: None  (b).Global H: No  (c).RPN: No attention 
% \end{tablenotes}
% \end{threeparttable}
% \end{center}
% \end{table}

\begin{table}[t]
\begin{center}
\begin{threeparttable}
\caption{Ablation Study}
\begin{tabular*}{\textwidth}{l @{\extracolsep{\fill}} lcccccc}
\hline
\textbf{Method} & \multicolumn{2}{l}{\textbf{Flickr30k Entities}} & \multicolumn{2}{l}{\textbf{Referlt Game}} & \multicolumn{2}{l}{\textbf{Visual Genome}} \\
& \textbf{R@1} & \textbf{mAP}    & \textbf{R@1}   & \textbf{mAP}    & \textbf{R@1}   & \textbf{mAP}    \\
\hline
\textbf{MAGNet}      & \textbf{60.20} & \textbf{0.4956} & \textbf{71.60} & \textbf{0.6052} & \textbf{28.85} & \textbf{0.1892} \\
MAGNet(a)            & 52.90 & 0.4293 & 69.95 & 0.6129 & 29.00 & 0.1823 \\
MAGNet(b)            & 49.65 & 0.3672 & 68.00 & 0.5505 & 26.50 & 0.1316 \\
MAGNet(c)            & 52.90 & 0.3730 & 68.95 & 0.5813 & 28.40 & 0.1144 \\
\hline
\end{tabular*}
\begin{tablenotes}
    \small
    \item MAGNet: Our model  (a) without word embedding (b) without global H (c) without attention-assisted RPN   
\end{tablenotes}
\end{threeparttable}
\end{center}
\end{table}

\section{Conclusion}
In this work, we utilize an encoder-decoder language model to fuse the input image and the natural language query and train an attention distribution over the input image which encodes both the local and global understanding of the query in relation to the input image. We utilize the generated context to train an attention-assisted region proposal network to generate proposals relevant to the query phrase and train an attention-assisted region CNN to classify these proposals in a Faster-RCNN framework. We call this framework the Multi-region Attention-assisted Grounding network (MAGNet). With this MAGNet framework, our model is independent of external proposal generation systems and without additional information it can develop understanding of the query phrase in relation to the image to achieve respectable results in Flickr30k entities and 12\% improvement over the state-of-the-art in ReferIt game. Additionally, our model is capable of grounding multiple regions for a query phrase, which is more suitable for real-life applications.  The use of attention distribution also makes the model more interpretable than other existing works.

\bibliographystyle{unsrt}

\clearpage

\end{document}